\documentclass[sigconf]{acmart}
\setcopyright{none}
\AtBeginDocument{%
  }

 \acmConference[KDD '24]{}{August 25--29,
   2024}{Barcelona, ESP}
\begin{document}

%%
%% The "title" command has an optional parameter,
%% allowing the author to define a "short title" to be used in page headers.
\title{A Scale-Invariant Diagnostic Approach Towards Understanding Dynamics of Deep Neural Networks }
%Non-Linear Connectionist Model 
%%
%% The "author" command and its associated commands are used to define
%% the authors and their affiliations.
%% Of note is the shared affiliation of the first two authors, and the
%% "authornote" and "authornotemark" commands
%% used to denote shared contribution to the research.
\author{Ambarish Moharil}
\authornotemark[1]
\affiliation{%
  \institution{Jhernonimus Academy of Data-Science,}
  \city{}
  \country{}
}
\affiliation{%
  \institution{Tilburg University}
  \city{}
  \country{}
}
\email{a.s.moharil@tilburguniversity.edu}

\author{Damian Tamburri}
\affiliation{%
  \institution{Eindhoven University of Technology}
  \city{}
  \country{}}
\email{d.a.tamburri@tue.nl}

\author{Indika Kumara}
\author{Willem-Jan Van Den Heuvel}
\affiliation{%
  \institution{Tilburg University}
  \city{}
  \country{}
}
\email{i.p.k.weerasingha.dewage@tue.nl}
\email{w.j.a.m.v.d.heuvel@jads.nl}

\author{Alireza Azarfar}
\affiliation{%
  \institution{ Deloitte Touche Tohmatsu Limited}
  \city{}
  \country{}}
\email{aazarfar@deloitte.nl}
%%
%% By default, the full list of authors will be used in the page
%% headers. Often, this list is too long, and will overlap
%% other information printed in the page headers. This command allows
%% the author to define a more concise list
%% of authors' names for this purpose.
\renewcommand{\shortauthors}{Moharil et al.}

%%
%% The abstract is a short summary of the work to be presented in the
%% article.
\begin{abstract}
% This paper presents a scale-invariant methodology employing Fractal Geometry to analyze and \textit{explain} the nonlinear dynamics of complex \textit{connectionist} systems during and after optimization. Our approach explores architectural self-similarity in Deep Neural Networks (DNNs), enabling multi-scale studies of network dynamics. We introduce a framework to quantify the fractal dimension of network segments, assessing their \textit{roughness} and enhancing understanding of their descriptive capabilities and temporal behavior. By integrating principles from Chaos Theory, we enhance the visualization of fractal segment evolution within an interpretable phase-flow domain, aiding in the identification and analysis of attractors and their characteristics. Furthermore, we outline a method to reconstruct the fractal representation of the network using a Graph-Based Neural Network, calculating non-linear relationships among network fragments via an exponential kernel. This generates an adjacency matrix for hypergraph construction, providing a foundational step toward comprehensive model diagnostics and advancing intrinsic explainability within connectionist AI.
This paper introduces a scale-invariant methodology employing \textit{Fractal Geometry} to analyze and explain the nonlinear dynamics of complex connectionist systems. By leveraging architectural self-similarity in Deep Neural Networks (DNNs), we quantify fractal dimensions and \textit{roughness} to deeply understand their dynamics and enhance the quality of \textit{intrinsic} explanations. Our approach integrates principles from Chaos Theory to improve visualizations of fractal evolution and utilizes a Graph-Based Neural Network for reconstructing network topology. This strategy aims at advancing the \textit{intrinsic} explainability of connectionist Artificial Intelligence (AI) systems.
\end{abstract}
\vspace{-1mm}
\maketitle
\section{Introduction}

Explainable Artificial Intelligence (XAI) seeks to demystify decision-making in complex Machine Learning and Deep Learning systems~\cite{islam2021explainable}. While there is no universal definition of explainability, Liao et al.~\cite{Liao_2020} describe it simply as \textit{"an answer to a question"}. Adopting \textit{connectionism} has greatly enhanced modeling capabilities regarding physical and informational complexities through complex non-linear dynamical systems of independently communicating units~\cite{Havel, Wang2017SymbolConnectionism}. These systems, often referred to as \textit{black-box} and partially \textit{chaotic}, show a sensitive dependence on initial conditions, complicating predictions about their long-term behavior~\cite{lorenz1972predictability}. Moreover, the inherent non-linearity across the network architecture suggests a connectionist self-symmetry, invariant across different scales of observation~\cite{fractal_review}.

This understanding is crucial for comprehending \textit{non-linearities} and \textit{emergence} in such networks. Traditional \textit{surrogate} methods like LIME, which offer post-hoc explanations via sparse linear feature representations, fail to capture the network's non-linear interactions and dynamic behaviors during optimization \cite{ribeiro2016why, islam2021explainable}. However, recent advancements in the use of fractal features for network analysis and the computation of fractal dimensions in complex systems suggest a new segmentation approach for creating a \textit{non-linear} connectionist representation across multiple scales~\cite{fractal, fractal_review}. Inspired by Mandelbrot’s work in fractal geometry and discoveries of non-linear attractor behaviors~\cite{attractor_training,mandelbrot1967how, feigenbaum1983universal}, our research leverages fractal analysis to delve deeply into connectionist network architectures. By evaluating fractal dimensions and roughness, we gain insights into network connectivity and emergent phenomena, thereby enhancing our understanding of system dynamics and aiding in the identification of cyclic \textit{attractors}, as demonstrated in Kauffman’s studies on Random Boolean Networks (RBNs)~\cite{kauffman1993origins}. Our approach aims to augment the \textit{intrinsic} explainability of connectionist networks and emergent phenomena by addressing two fundamental questions: \textbf{RQ1.} \textit{"How can we create a non-linear connectionist representation across multiple scales for DNNs?"} and \textbf{RQ2.} \textit{"To what extent does such a representation enhance the explainability of nonlinear interactions in DNNs?"}

\section{Proposed Methodology}
Our methodology segments the network at specific scales during and after optimization, generating a fractal representation of network connections using a graph-based surrogate.
Segmenting the layers of a network (DNN), in turn, implies partitioning the associated parameter matrices, as it is the parameters that \textit{form} and \textit{deform} the network.
Formally, consider a connectionist network \( f \) with \( L_p \) layers. For any given layer \( L_i \), connected to subsequent layer \( L_j \) where \( i \neq j \), the parameter matrix \( W_{n \times m} \) represents the connections, where \( n \) and \( m \) denote the number of nodes in layers \( L_i \) and \( L_j \) respectively. The fractal dimension \( FD \) of this matrix is calculated using the box-counting method \cite{fractal}, defined as:
$ FD = \frac{\ln(N)}{\ln(1/r)} \in \mathbb{R} $
where \( N \) is the number of \( r \times r \) boxes needed to cover the matrix. The segment size \( r \), crucial for fractal analysis, must satisfy: $ r > 1 $, to avoid granularity at the level of individual matrix elements, and: $ r < \min(n, m) $, to prevent oversimplification by covering the matrix with a single box. The specific ranges for \( r \) are defined as follows:
$$
r \in \begin{cases} 
    [2, \lfloor \frac{n + 1}{2} \rfloor], & \text{if } n = m \text{ and } n \hspace{-2mm} \mod2 \neq 0\\
    [2, \frac{n}{2}], & \text{if } n = m \text{ and } n\hspace{-2mm}  \mod 2 = 0 \\
    [2, \lfloor \frac{\min(n, m) + 1}{2} \rfloor], & \text{if } n \neq m \text{ and } \min(n, m)\hspace{-2mm} \mod2 \neq 0 \\
    [2, \frac{\min(n, m)}{2}], & \text{if } n \neq m \text{ and } \min(n, m) \hspace{-2mm} \mod 2 = 0
\end{cases}
$$
% To segment the matrix \( W_{nm} \), we proceed by dividing it into smaller segments of size \( r \times r \). The segmentation process iterates over the matrix dimensions and extracts sub-matrices according to the calculated \( r \). For each valid starting point \( (i, j) \) within the matrix boundaries that does not exceed the matrix dimensions \( n \) and \( m \), segments are defined by:
% $$ S_{ij} = W_{nm}[i \cdot r: (i+1) \cdot r, j \cdot r: (j+1) \cdot r] $$
% for \( i = 0 \) to \( \left\lceil \frac{n}{r} \right\rceil - 1 \) and \( j = 0 \) to \( \left\lceil \frac{m}{r} \right\rceil - 1 \).
% For square matrices where \( n = m \):
% \begin{itemize}
%     \item If \( n \) is odd, the segmentation may result in edge overlapping due to the stride being set to 1 to accommodate the central segment.
%     \item If \( n \) is even, the segmentation results in complete non-overlapping matrices, as the dimensions perfectly accommodate an integer number of \( r \times r \) segments without requiring a stride adjustment.
% \end{itemize}
\begin{figure}[h!]
    \centering
    \includegraphics[width = 0.9\linewidth]{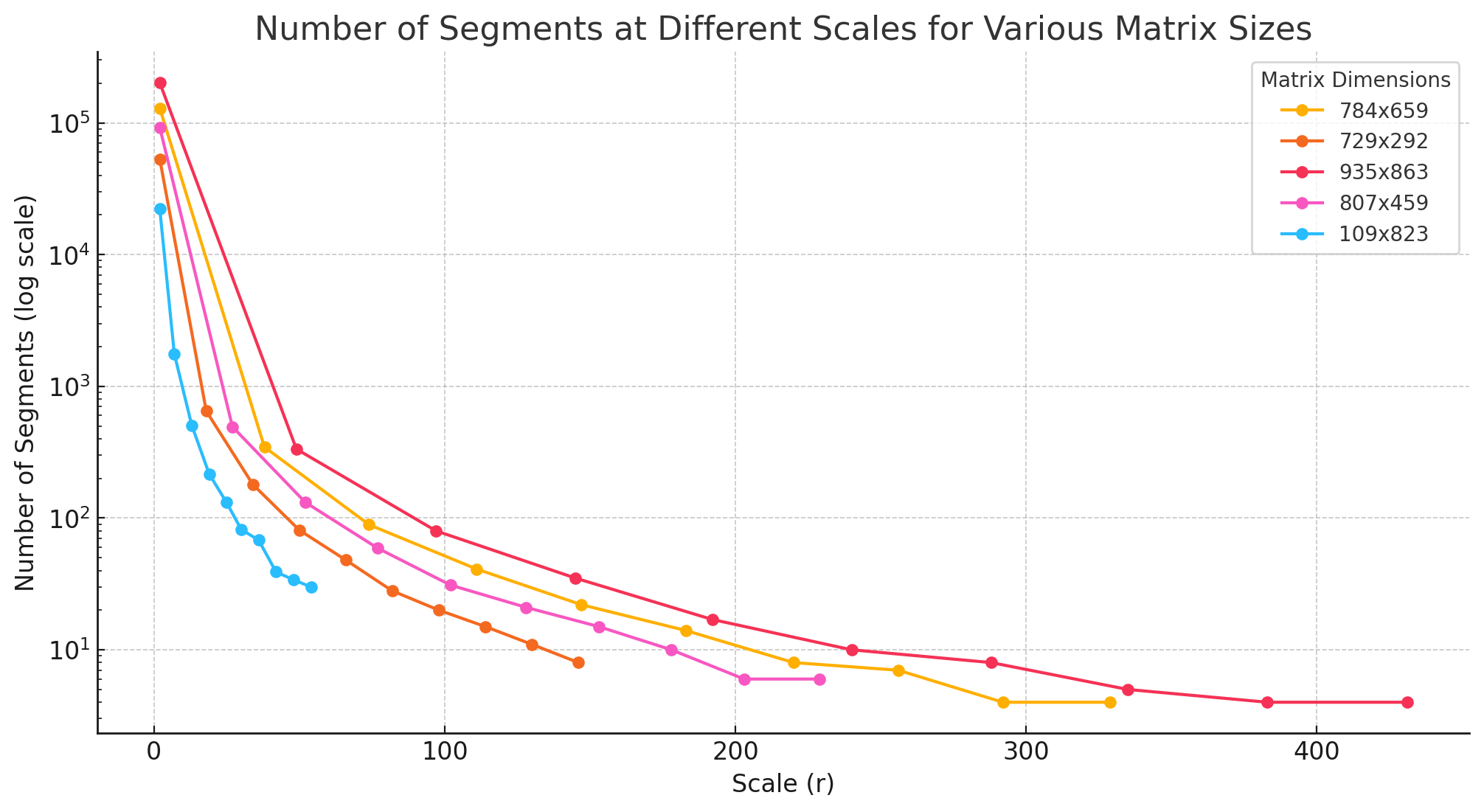}
    \caption{Visualizing the $\#$segments (log scale) at various \textit{valid} scales $r$ for different matrix dimensions $n \times m$.}
    \vspace{-4mm}
    \label{fig:segment_size}
\end{figure}
The segmentation extracts sub-matrices based on \( r \), iterating over dimensions \( n \) and \( m \) where $ n , m > 2 $ as (Fig \ref{fig:segment_size}):
\begin{equation}
     S_{ij} = W_{nm}[i \cdot r : (i+1) \cdot r, j \cdot r : (j+1) \cdot r] 
\end{equation}
for \( i = 0 \) to \( \left\lceil \frac{n}{r} \right\rceil - 1 \) and \( j = 0 \) to \( \left\lceil \frac{m}{r} \right\rceil - 1 \).
\footnote{Segmentation overlap depends on the parity of matrix dimensions \( n \) and \( m \) in \( W_{nm} \). For square matrices (\( n = m \)), an odd \( n \) means a stride of 1, causing edge overlaps, while an even \( n \) allows perfect tiling without overlaps. For non-square matrices, odd \(\min(n, m)\) results in overlaps for coverage, whereas even \(\min(n, m)\) ensures perfect alignment without overlaps.}
% \footnote{Overlap conditions in the segmentation process are dependent on the parity of the dimensions \( n \) and \( m \) of the matrix \( W_{nm} \). For square matrices, where \( n = m \), the stride is set to 1 if \( n \) is odd, leading to some edge overlaps to accommodate the center of the matrix. Conversely, if \( n \) is even, the segments fit perfectly without any overlap, resulting in a complete tiling of the matrix. In the case of non-square matrices, if the smaller dimension, \( \min(n, m) \), is odd, overlaps occur in this dimension to ensure coverage; however, if the smaller dimension is even, the segments again align perfectly without any overlap.} 
This adaptive approach to segmentation ensures optimal coverage and granularity for analytical purposes, accounting for structural variations in the matrix dimensions. % For each segment \( S_{ij} \) of the network matrix \( W_{nm} \), obtained by segmenting at scale \( r \), the Fractal Dimension Vector (FDV) is computed across a set of valid scales \( Q \). The FDV is determined by the product of the fractal dimension of \( S_{ij} \) at each scale \( r_q \) and the entropy of \( S_{ij} \), reflecting the segment's complexity and information content. Mathematically, this is expressed as:
% \begin{equation}
%      FDV(S_{ij}) = \{ FD(S_{ij}, r_q) \times H(S_{ij}) \, | \, r_q \in Q \} 
% \end{equation}
% where \( FD(S_{ij}, r_q) \) represents the fractal dimension of segment \( S_{ij} \) at scale \( r_q \), and \( H(S_{ij}) \) is the entropy of the segment, calculated as:$ H(S_{ij}) = -\sum p(x) \log p(x) $ with \( p(x) \) being the probability distribution of the elements within \( S_{ij} \). This vector encapsulates the roughness and entropy of each segment across scales. 
We furthermore use the activation map $A_{g,h}^{i}, \backepsilon g,h \in \mathbb{R}$, of layer $L_{i}$ to study the non-linear interactions between fractal segments of subsequent layers, computed at scale $r_{q}$. Given two fractal segments $S_{\omega}$ and $S_{\lambda}$, we apply the exponential kernel taking inspiration from~\cite{ribeiro2016why, WANG2023126651}, computing the edge values between segments across layers $L_{i}:L_{p}$ accounting for the local influence of the parameter segments, as follows :
\begin{equation}
    e_{\omega, \lambda} = \gamma \cdot exp(||\alpha_{\omega} - \alpha_{\lambda}||)
\end{equation}
Where $\gamma$ is the spread of the kernel, $\alpha_{\omega}$ and $\alpha_{\lambda}$ and are segment specific features at scale $r_{q}$, computed as :
\begin{equation}
    \alpha_{S_{i,j}} = A_{g,h}^{'i} \cdot FD(S_{i,j}, r_{q}) \cdot H(S_{i,j})
\end{equation} , 
% where $A_{g,h}^{'i}$ is the flattened feature map of $A_{g,h}^{i}$ and \( H(S_{ij}) \) is the entropy of the segment, calculated as $ H(S_{ij}) = -\sum p(x) \log p(x) $ with \( p(x) \) being the probability distribution of the elements within \( S_{ij} \). Having extracted the non-linear relationships amongst fractal segments in the form of an adjacency matrix $\mathcal{A}$, we propose learning the network representation at a given scale $r_{q}$ using a Graph-Based Neural Network taking inspiration from Wang et al. \cite{WANG2023126651} for each instance within the dataset and aggregating the learned graphs in a resulting fractal hypergraph.  Exploration of the underlying structural space ($\mathcal{A}$) and learning of the information within the space can be achieved, resulting in the following outcome:
Where \(A_{g,h}^{'i}\) is the flattened feature map of \(A_{g,h}^{i}\) and \(H(S_{ij})\) is the entropy of the segment, calculated as \(H(S_{ij}) = -\sum p(x) \log p(x)\) with \(p(x)\) representing the probability distribution within \(S_{ij}\). Following the extraction of non-linear relationships among fractal segments, represented as an adjacency matrix \(\mathcal{A}\), we utilize a Graph-Based Neural Network to learn the network representation at scale \(r_{q}\), inspired by Wang et al.~\cite{WANG2023126651}. This approach allows for the exploration and learning of structural space (\(\mathcal{A}\)), culminating in the aggregation of learned graphs into a fractal hypergraph and achieving comprehensive insights into the underlying informational dynamics as:
\begin{equation}
\Omega(A_{g,h}^{(i,k)}, \mathcal{A}) = \sigma\left(\hat{D}^{-\frac{1}{2}} \hat{\mathcal{A}} \hat{D}^{\frac{1}{2}} A_{g,h}^{(i,k)} S_{i,j}^{(k)}\right)
\end{equation}
where $\sigma$ is the activation function, $A_{g,h}^{(i,k)}$ the feature map of $k^{th}$ segment of $i^{th}$ layer along with the parameter segment $S_{i,j}^{k}$ and $\hat{D}$ is the diagonal degree of $\mathcal{A}$.
\vspace{-2mm}

\section{Preliminary Results}
\begin{figure}[ht]
    \centering
    % First Image
    \begin{minipage}[b]{0.495\linewidth}
        \centering
        \includegraphics[width=\linewidth]{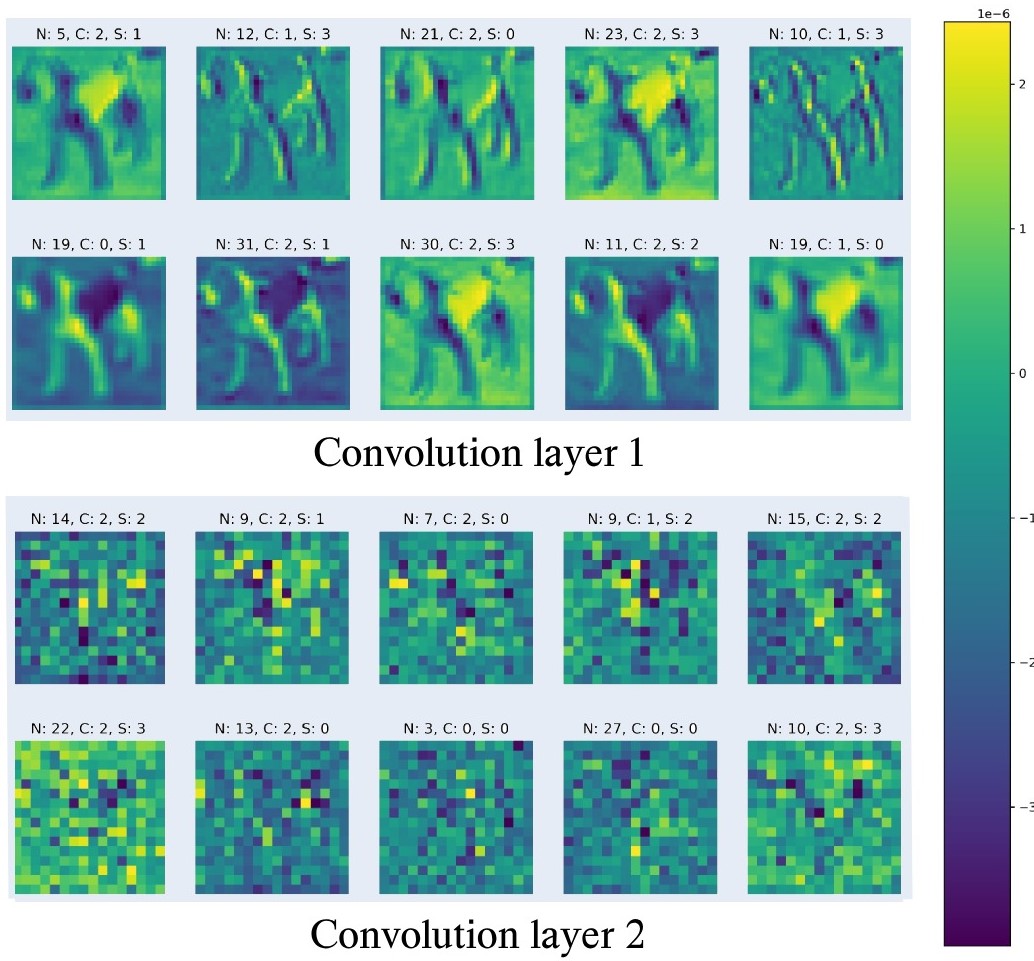}
        \vspace{-4mm} % Adjust vertical spacing as necessary
    \end{minipage}
    % Second Image, no horizontal fill for minimal spacing
    \begin{minipage}[b]{0.495\linewidth}
        \centering
        \includegraphics[width=\linewidth]{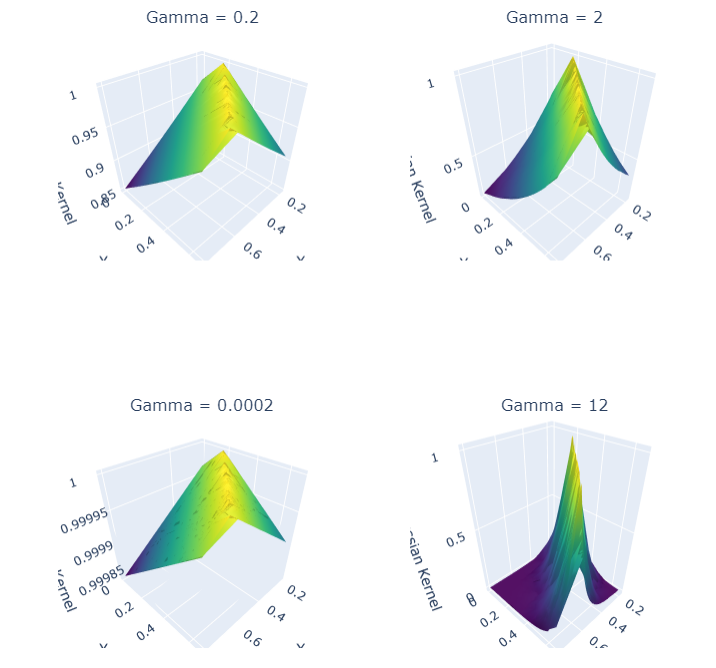}
        \vspace{-4mm} % Adjust vertical spacing as necessary
    \end{minipage}
    \caption{\small Left image shows features learned by neuron-channels across two convolution layers. Right image visualizes the exponential kernel interactions between fractal segment features for various $\gamma$ values.}
    \vspace{-5mm}
    \label{fig:activations} % This label now applies to the entire figure environment
\end{figure}

\begin{figure}[h!]
    \centering
    % First Image
    \begin{minipage}[b]{0.5\linewidth}
        \centering
        \includegraphics[width=\linewidth]{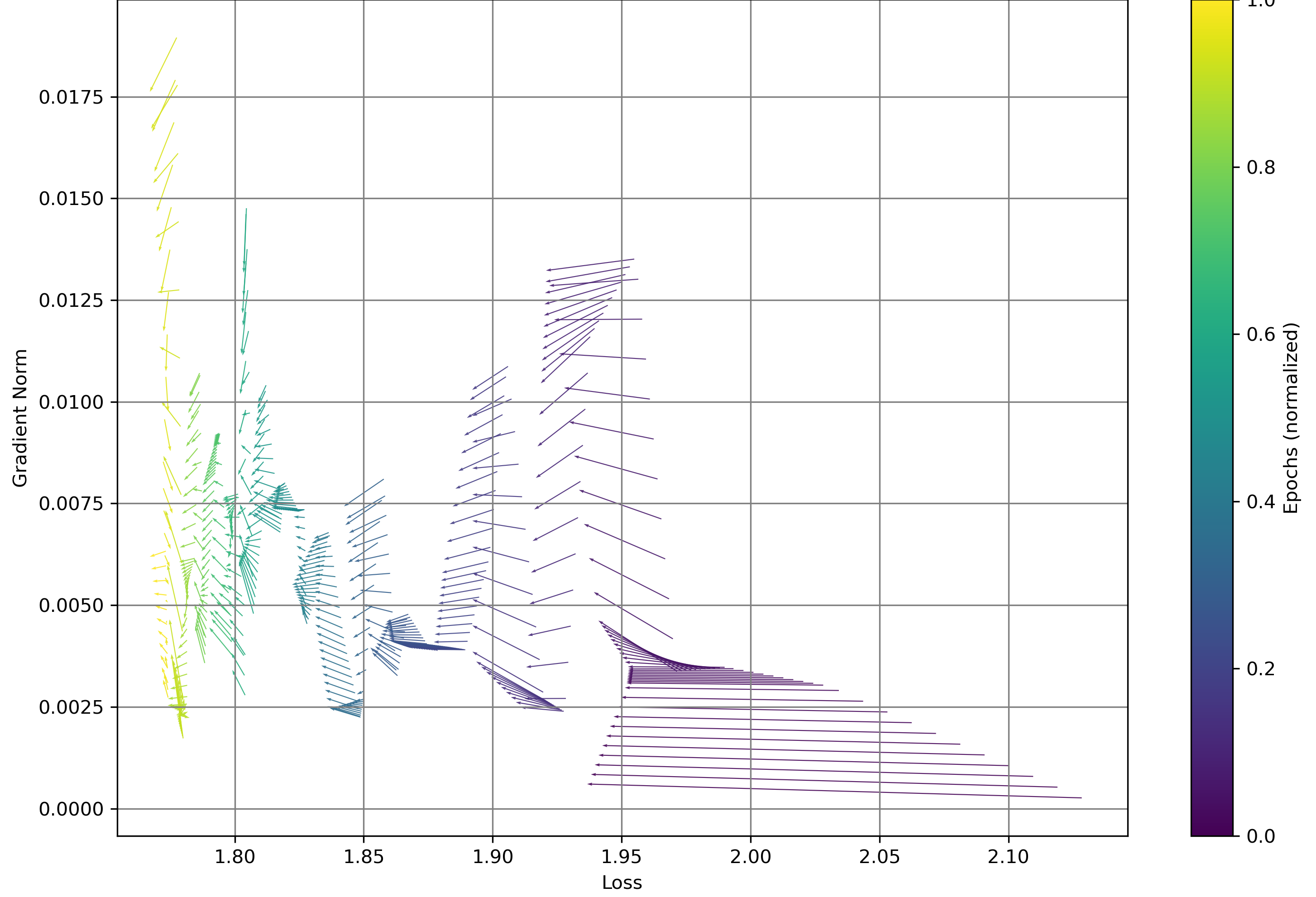}
        \vspace{-0.7mm}
        \label{fig:pha1}
    \end{minipage}
    \hfill % Optional: add some horizontal spacing
    % Second Image
    \vspace{-8mm}
    \begin{minipage}[b]{0.48\linewidth}
        \centering
    \includegraphics[width=\linewidth]{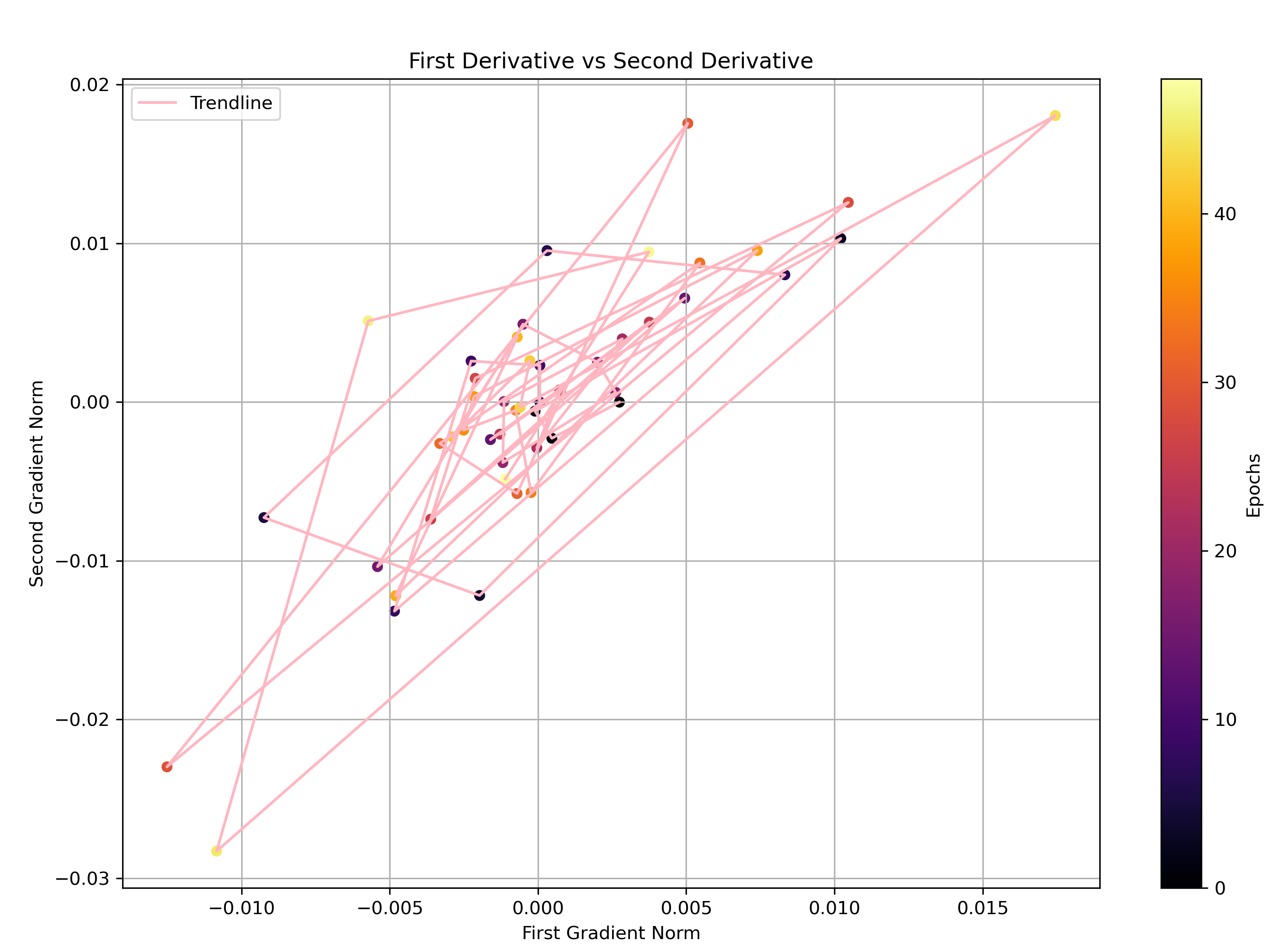}
        \vspace{-0.5mm}
        \label{fig:pha2}
    \end{minipage}
    \vspace{-2mm}
    \caption{Phase-Flow Diagrams}
    \label{fig:phaseflow-x}
\end{figure}

In our study, we examine multiclass classification on the MNIST dataset~\cite{deng2012mnist} using a CNN with two convolutional layers of 32 and 64 neurons and two fully connected layers, each convolutional layer utilizing a kernel size of 3 with padding and stride set to 1. We train the model over 50 epochs with a learning rate of $6e-4$ using the Adam Optimizer, and analyze gradients and loss post each epoch through a segmentation of model weights, as described in equation $1$. The weights of the first convolutional layer are a $4-D$ tensor $[32,3,3,3]$ and the second layer $[64,32,3,3]$, segmented into four overlapping segments per neuron, scaled by $r=2$. We evaluate segment features to determine their influence on inputs as detailed in equation $3$, with Figure \ref{fig:activations} (left) illustrating the captured segment features, and Figure \ref{fig:activations} (right) displaying the exponential kernel interactions between segments, indicating local influence across layers. We analyze the phase flow graph $\forall_{j=1}^{M}\forall_{i=1}^{Q} \frac{\partial\mathcal{L}}{\partial(W_{s_{i}}^{j})} \text{ Vs } \mathcal{L}$ for a network with \(M\) layers and \(Q\) segments per layer, where \(\mathcal{L}\) is the cross-entropy loss. Figure \ref{fig:phaseflow-x} (left) depicts the learning trajectory of a segment, highlighting an initial acceleration followed by a gradual deceleration, and Figure \ref{fig:phaseflow-x} (right) shows a plot of \(\frac{\partial}{\partial t}\) vs \(\frac{\partial^2}{\partial t^2}\), evidencing a positive trend. The epochs, color-coded, reveal a decreasing trend in gradient norms, suggesting model stabilization. Data convergence in later epochs toward the center indicates the formation of an attractor, enhancing predictability and training stability.
\vspace{-1mm}

% We analyze a phase flow graph $\forall_{j=1}^{M}\forall_{i=1}^{Q} \frac{\partial\mathcal{L}}{\partial(W_{s_{i}}^{j})} \text{ Vs } \mathcal{L}$ across a network with M layers and Q segments per layer, where $\mathcal{L}$ denotes the cross-entropy loss function, to visualize the learning behavior of individual fractal segments. On the one hand, Figure \ref{fig:phaseflow} (left) depicts the learning trajectory of a segment, showing an initial rapid acceleration followed by a gradual deceleration, similar to a vehicle slowing down. On the other hand, Figure \ref{fig:phaseflow} (right) displays a plot of \(\frac{\partial}{\partial t}\) vs \(\frac{\partial^2}{\partial t^2}\), showing a positive correlation with an upward trendline. The epochs, color-coded, show a decreasing trend in gradient norms, indicating model stabilization. Variability and overlap in the data highlight sensitivity to initial conditions and chaotic dynamics. The convergence of data points in later epochs toward the center suggests the formation of an attractor, marking increased predictability and stability in the training process.
\vspace{-4mm}
\section{Research Roadmap}
Our future research in Explainable Artificial Intelligence (XAI) encompasses three interconnected strategic categories, aiming to deepen the integration and sophistication of graphical models and analytical techniques. In \textit{Enhancements in Graph-Based Surrogates}, we focus on advancing fractal feature learning using graph-based surrogates, improving model capabilities through node embeddings and neural message passing~\cite{kipf2017semisupervised}, and extending our approaches to incorporate semantic feature analysis across various DNN architectures using saliency models~\cite{hou2023decoding}. The \textit{Advanced Visualization and Theoretical Approaches} involve applying renormalization theory for efficient feature distillation~\cite{erbin2022renormalization} and exploring attractor behaviors (fixed, periodic, quasi, aperiodic) \cite{attractor_training, Chatterjee1992} within network segments along with identifying \textit{answerable} questions regarding system dynamics~\cite{Liao_2020}. Finally, \textit{Comprehensive System Analysis} leverages Automated Machine Learning (AutoML) for hyperparameter sampling to facilitate a robust analysis of system behaviors as either \textit{dissipative} or \textit{conservative}~\cite{Chatterjee1992}, enhancing the intrinsic explainability of AI models~\cite{zöller2021benchmark, samek2017explainable}. Together, these efforts aim to advance the transparency and interpretability of AI systems within the field of XAI, using the concepts from non-linear system dynamics.
\bibliographystyle{unsrt}
\bibliography{main}

%%
%% If your work has an appendix, this is the place to put it.

\end{document}